\documentclass{article}

\PassOptionsToPackage{numbers, compress}{natbib}
\usepackage[final]{neurips_2024}
\usepackage[resetlabels,labeled]{multibib}

\usepackage[utf8]{inputenc} % allow utf-8 input
\usepackage[T1]{fontenc}    % use 8-bit T1 fonts
\usepackage{hyperref}       % hyperlinks
\usepackage{url}            % simple URL typesetting
\usepackage{amsfonts}       % blackboard math symbols
\usepackage{nicefrac}       % compact symbols for 1/2, etc.
\usepackage{microtype}      % microtypography
\usepackage{xcolor}         % colors
\usepackage{enumitem}
\usepackage{graphicx}
\usepackage{wrapfig}
\usepackage{multirow}
\usepackage{makecell}
\usepackage{caption}
\usepackage{algorithmicx}  
\usepackage{algorithm}
\usepackage{algpseudocode}  
\usepackage{bm}
\usepackage{arydshln}
\usepackage{float}
\usepackage{subcaption}
\usepackage{amsmath}
\usepackage{booktabs}
\usepackage{arydshln}
\usepackage{amsmath}
\definecolor{mydarkgreen}{RGB}{0,115,0}

\title{Voxel Mamba: Group-Free State Space Models for Point Cloud based 3D Object Detection}

\author{%
  Guowen Zhang$^{1,2}$, Lue Fan$^3$, Chenhang He$^1$, Zhen Lei$^{2,3}$, Zhaoxiang Zhang$^{2,3,*}$, Lei Zhang$^{1, }\thanks{Corresponding Author}$\\
  $^1$The Hong Kong Polytechnic University\\
  $^2$Center for Artificial Intelligence and Robotics, HKISI, CAS\\
  $^3$Institute of Automation, Chinese Academy of Sciences\\
  \texttt{guowen.zhang@connect.polyu.hk, \{csche, cslzhang\}@comp.polyu.edu.hk}\\
  \texttt{\{fanlue2019, zlei, zhaoxiang.zhang\}@ia.ac.cn}\\
  Code: {\color{red}{\href{https://github.com/gwenzhang/Voxel-Mamba}{https://github.com/gwenzhang/Voxel-Mamba}}}
  }

\begin{document}

\maketitle

\begin{abstract}

Serialization-based methods, which serialize the 3D voxels and group them into multiple sequences before inputting to Transformers, have demonstrated their effectiveness in 3D object detection. 
However, serializing 3D voxels into 1D sequences will inevitably sacrifice the voxel spatial proximity. Such an issue is hard to be addressed by enlarging the group size with existing serialization-based methods due to the quadratic complexity of Transformers with feature sizes.
% Furthermore, restricted by quadratic complexity, Voxel Transformers always group voxels into fixed-length subsets to balance efficiency with narrow receptive fields.
% while Transformers achieve remarkable performance, it is restricted by their quadratic complexity with respect to the sequence length. 
% Thus, Voxel Transformers always group voxels into small sets to balance efficiency but limit modeling capability with narrow receptive fields.
% It is highly anticipated that we can develop a new architecture to possess both high modeling capability and computational efficiency.
Inspired by the recent advances of state space models (SSMs), we present a Voxel SSM, termed as Voxel Mamba, which employs a group-free strategy to serialize the whole space of voxels into a single sequence.
The linear complexity of SSMs encourages our group-free design, alleviating the loss of spatial proximity of voxels.
% model long-range dependencies with larger receptive fields. 
% Voxel Mamba adopts a group-free strategy to serialize voxels into one single sequence.
% It can avoid inefficient grouping operations and prevent limitations on the receptive field.
% provide each voxel with a global receptive field.
To further enhance the spatial proximity, we propose a Dual-scale SSM Block to establish a hierarchical structure, enabling a larger receptive field in the 1D serialization curve, as well as more complete local regions in 3D space.
Moreover, we implicitly apply window partition under the group-free framework by positional encoding, which further enhances spatial proximity by encoding voxel positional information. 
% In order to effectively preserve the locality in sequences, we serialize scene voxels through space-filling curves and offer rich 3D position information by a simple yet powerful implicit window partition.
% To enhance voxel proximity in sequences, we propose Asymmetrical State Space Models (ASSMs) to enlarge the effective receptive fields along curves and establish the relation between voxels that exhibit close spatial proximity in 3D space.
% Voxel Mamba achieves not only higher accuracy than the well-established Voxel Transformers (\textit{e.g.}, DSVT) and SpCNN (\textit{e.g.}, PV-RCNN) but also shows significant advancements in computational and memory efficiency, as demonstrated on the Waymo Open dataset and nuScenes dataset. Codes will be made publicly available.
Our experiments on  Waymo Open Dataset and nuScenes dataset show that Voxel Mamba not only achieves higher accuracy than state-of-the-art methods, but also demonstrates significant advantages in computational efficiency.
% higher accuracy than the  Transformer-based or Sparse C (\textit{e.g.}, DSVT) and SpCNN (\textit{e.g.}, PV-RCNN) but also shows significant advancements in computational and memory efficiency, as demonstrated on the Waymo Open dataset and nuScenes dataset. Codes will be made publicly available.

\end{abstract}

\label{sec:abstract}

\section{Introduction}

\href{}{}
% Learning effective representation from point clouds plays a crucial role in 3D object detection.
% The key challenge for point cloud understanding is that point clouds are sparse, uneven, and irregular. 
% To handle those challenges, previous works mainly focus on designing dedicated sparse operations such as sparse convolution~\cite{Second,PVRCNN} and Transformers~\cite{SST,DSVT}. In this paper, we propose Voxel Mamba, an effective State Space Models (SSMs)-based backbone to overcome limitations of previous works and further enhance detection performance. 

\begin{figure}[t]
\centering
\includegraphics[width=0.8\textwidth]{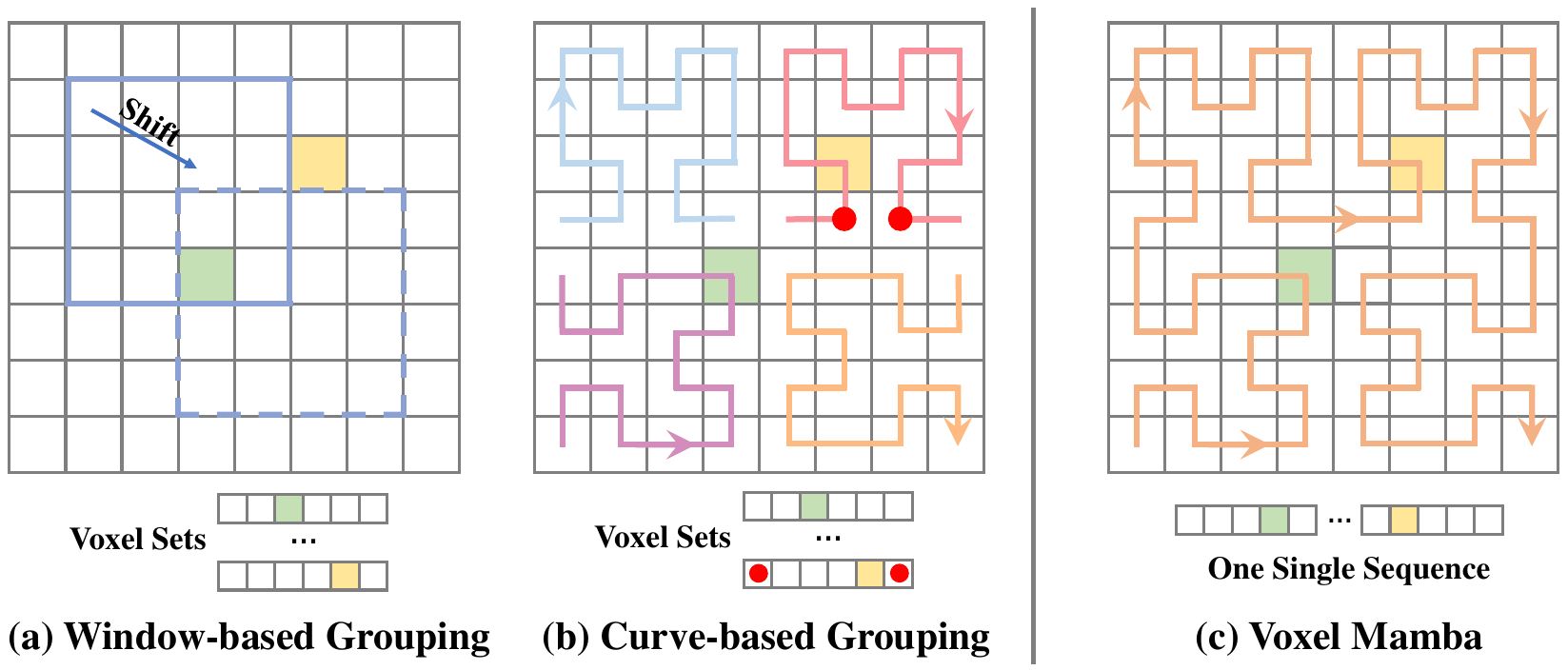}
\caption{Comparison between (a) window-based grouping, (b) curve-based grouping, and (c) our proposed single group modeling by Voxel Mamba.}
\label{Fig::Insight}
\vspace{-4mm}
\end{figure}

% for SpCNN
LiDAR-based 3D object detection from point clouds plays an important role in applications of autonomous driving~\cite{Kitti,e2ead_survey}, virtual reality~\cite{VR_3Ddetection}, and robots~\cite{robot}. The sparsely, unevenly and irregularly distributed point cloud data make the efficient and effective 3D object detection a very challenging task. To address these long-standing challenges, researchers have recently proposed several strategies to improve the model architecture. One strategy is to switch from PointNet-based models~\cite{Frustumpointnet,Voxelnet,Pointrcnn} to sparse convolutional neural network (SpCNN)-based models~\cite{Second,PVRCNN,VoxelRCNN,FSD,PVRCNN++,SA_SSD} in order for more effective feature extraction.
% SpCNN-based
% However, the small receptive fields limit the representation capacity of sparse convolutions, as highlighted by previous arts~\cite{VoxelTransformer,SST}.
However, the sparse convolution is unfriendly for deployment and optimization, requiring tremendous engineering efforts.
% for group-based Transformers 
% Second, Voxel Transformers achieved superior performance and demonstrated their potential as alternatives to SpCNN with a larger receptive field and dynamic weights. 
% Second, Voxel Transformers have recently become an alternative for SpCNN due to their better capability to model long dependencies.
% achieved superior performance with a larger receptive field and dynamic weights. 
% Inspired by Vision Transformers~\cite{swin_transformer, ViT}, various attention-based methods have been introduced to point cloud processing.
% Second, Voxel Transformers achieved superior performance and demonstrated their potential as alternatives to SpCNN.
Therefore, another strategy is to switch from SpCNN to serialization-based Transformers to address this issue~\cite{SST,DSVT,Flatformer,Ptv3,VoxelTransformer}. 
% often involve serializing 3D spatial voxels into sequences to apply standard Transformers, which inevitably decreases voxel proximity.
These methods usually group non-empty 3D voxels into multiple short sequences by serialization techniques such as window partition~\cite{DSVT,SST,Flatformer}, Z-shape sorting~\cite{Octformer}, and Hilbert sorting~\cite{Ptv3}, as shown in Figs.~\ref{Fig::Insight} (a) and (b), where a sequence is a group of voxels to be processed by Transformer layers.
% Furthermore, the sparsity of 3D points and the varying number of non-empty voxels in local regions present challenges for processing variable-length sequences with Transformers.

However, the serialization of voxels will inevitably sacrifice their spatial proximity. Some neighboring voxels can be far apart from each other after serialization, as illustrated by the two red points in Fig.~\ref{Fig::Insight} (b).
Such a loss of proximity is difficult to be addressed in the existing serialization methods~\cite{Ptv3,Octformer,Flatformer,DSVT,SST} because the group size is limited by the quadratic complexity of Transformers. This issue becomes even worse when neighboring voxels are grouped into different groups. 
% that can abandon grouping operations and provide each voxel with data-dependent global contexts. 
Inspired by the recent success of State Space Models (SSMs)~\cite{S4,Mamba,S5,H3,Vision_Mamba,Vmamba} in language and vision, in this work we propose a simple yet effective group-free design to address the loss of proximity. Specifically, we introduce a Voxel SSM, termed as \textit{Voxel Mamba}, for 3D object detection from point cloud.
The linear computational complexity of SSMs makes it feasible to treat all voxels as a single group and sort them into a single sequence. This results in a group-free modeling of voxels, which is more efficient and deployment-friendly than previous methods since no padding tokens are needed.
Nonetheless, even we can sort all voxels into a group-free sequence, it cannot be ensured that all of them are within an effective receptive field.  

To enhance the spatial proximity of Voxel Mamba, we further propose two modules with it. The first is the \textbf{D}ual-scale \textbf{S}SM \textbf{B}lock (DSB) by introducing the downsampling operations in SSMs. In specific, the forward SSM branches process the high-resolution voxel features, while the backward branches extract features from the low-resolution representation.
In this way, we integrate the hierarchical design with the bidirectional design in a more economical way.
More importantly, the hierarchy brings a larger effective receptive field for the serialized sequence so that the spatial proximity in local 3D regions can be enhanced.
The second module we introduced is the Implicit Window Partition (IWP). 
The window partition is a widely used strategy in previous methods~\cite{SST,DSVT} to enhance the proximity of voxels inside a window.
However, it impedes the proximity of voxels across windows and contradicts with our group-free principle.
We therefore propose an implicit window partition scheme to embrace its strengths while discarding its weaknesses.
In specific, we encode the voxel positions inside and across windows into embeddings for feature learning without explicitly conducting spatial window partition.
In this way, better voxel proximity can be achieved under our group-free design with minimal computational cost.

Our contributions are summarized as follows:
\begin{itemize}
    \item We propose Voxel Mamba, a group-free backbone for voxel-based 3D detection. Voxel Mamba abandons the grouping operation and serializes voxels into one single sequence, enabling better efficiency.
    % It can provide each voxel with global context through State Space Models.
    \item To mitigate the loss of spatial proximity due to serialization, we propose the Dual-scale SSM Block (DSB) and the Implicit Window Partition (IWP) to enhance the spatial proximity preservation of Voxel Mamba.
    % to combine the advantage of single stride architecture and multi-scale semantic features, which aggregates richer context information from the scene across various scales.
    \item Our method achieves superior performance to previous state-of-the-art methods on the large-scale Waymo Open dataset~\cite{Waymo} and nuScenes~\cite{Nuscenes} datasets.
\end{itemize}

% traverses each voxel based on Hilbert curve order without repetition. 
% We rearrange the voxels in
% \begin{table*}
% \centering
% \caption{The performance on the validation set of NuScenes.}
%  \vspace{-2mm}
% \resizebox{1\columnwidth}{!}{
% \begin{tabular}{c|c|c|c|c|c|c|c|c|c|c|c|c}
% 	\toprule[1pt]
%  \rowcolor{gray!20}
% Method & NDS & mAP & Car & Truck & Bus & T.L. & C.V. & Ped. & M.T. & Bike & T.C. & B.R. \\ 
% \hline
% DSVT~\cite{DSVT} & {71.1} & 66.4 & {87.4} & {62.6} & 75.9 & 42.1 & 25.3 & {88.2} & 74.8 & 58.7 & 77.8 & \textbf{70.9} \\
% \hline
% % \textbf{Voxel Mamba} (pillar) & 70.5 & \textbf{66.9} & 87.1 & 60.4 & \textbf{77.8} & \textbf{48.7} & \textbf{28.9} & 87.7 & \textbf{76.2} & \textbf{59.8} & 76.1 & 66.5 \\ 	
% \textbf{BiMamba} & 70.0 & {64.7} & 86.7 & 62.0 & {73.9} & {41.8} & {24.4} & 87.1 & {71.4} & {55.4} & 76.5 & 67.9 \\ 
% \toprule[1pt]
% \end{tabular}}
% \label{tbl:nuscenes_intro}
% \end{table*}

\label{sec:intro}

\section{Related Work}
\textbf{3D Object Detection from Point Clouds.} There are two major point cloud representations for 3D object detection, \textit{i.e.}, point-based and voxel-based ones. As in PointNet~\cite{Pointnet,Pointnetpp}, point-based methods~\cite{Frustumpointnet,Votenet,Pointrcnn,Fcaf3d,Groupfree} directly extract geometric features from small regions of raw points. However, those methods suffer from low inference efficiency and limited context features. Voxel-based methods~\cite{PVRCNN,PVRCNN++,Second,SA_SSD,Voxelnet,fsdv2,fan2023once,SuperSparse} convert raw points into regular grids through voxelization and then process them with sparse convolution~\cite{Second} or Transformers~\cite{SST,Voxset,DSVT}. 
Voxel-based methods are currently the main stream for 3D object detection. In terms of model architecture, voxel-based methods can be categorized into two groups, \textit{i.e.}, SpCNN-based~\cite{Second,Voxelnet,PVRCNN,PVRCNN++,VoxelNext,VoxelRCNN} and Transformers-based~\cite{DSVT,SST,Voxset,VoxelTransformer,Flatformer} ones. 
%
% Convolution-based methods~\cite{Second,Voxelnet,PVRCNN,PVRCNN++,VoxelNext,VoxelRCNN} utilize the 3D sparse CNNs which consist of submanifold and regular sparse convolution with small kernels to extract features. Motivated by the success of Transformers in computer vision, many researchers~\cite{DSVT,SST,Voxset,VoxelTransformer,Flatformer} group voxels in batches and introduce Transformers to capture long-range dependencies. 
%
Limited by the high computation complexity, SpCNN-based methods can only use small convolution kernels with restricted receptive fields, and Transformer based methods can only employ a small number of voxels in each group. In contrast, our proposed Voxel Mamba can capture long-range dependencies within the entire sequence while achieving faster inference speed than existing state-of-the-art methods.

\textbf{State Space Models.} Inspired by the continuous state space models (SSMs) in control systems, researchers~\cite{H3,S4,S5,Mamba} have introduced the SSMs into deep neural networks as a novel alternative to CNNs and Transformers. LSSLs~\cite{LSSL} adopts a simple sequence-to-sequence transformation, demonstrating the potential of SSMs. S4~\cite{S4} introduces a new parameterization method to SSMs to reduce the computation and memory cost. S5~\cite{S5} employs MIMO SSMs and perform efficient parallel scans based on S4. 
% GSS~\cite{Gated_ssms} designs a gated state space model and achieves a speedup of 2-3 times compared to S4. 
More recently, Mamba~\cite{Mamba} introduces input-dependent SSMs and builds a generic backbone, which is fairly competitive with the well-tuned Transformers. Vision Mamba~\cite{Vision_Mamba} employs bidirectional SSMs and position embedding to learn global visual context for vision tasks. Vmamba~\cite{Vmamba} employs a 2D-selective-scan to bridge the gap between 1D scanning and 2D plain traversing. 
PointMamba~\cite{pointmamba} is a pioneering work to leverage SSM for point cloud analysis, achieving impressive performance in point cloud object understanding.
Subsequently, many SSM-based methods~\cite{mamba3d,mamba4d,3dmambaipf,PCM,pointramba} are introduced for point cloud processing.
In this paper, we investigate the utilization of SSMs to establish a straightforward yet robust baseline for LiDAR-based 3D object detection in driving scenes.

\textbf{Space-filling Curve.} The space-filling curve~\cite{Spacefillingcurve} is a series of fractal curves that can go through each point in a multi-dimensional space without repetition. 
% It is widely utilized in databases~\cite{databases_1}, GIS~\cite{GIS}, and image compression~\cite{image_compession}. 
The classical space-filling curve includes Hilbert curve~\cite{hilbertcurve}, Z-order curve~\cite{zorder_curve}, and sweep curve, \textit{etc}. Those methods can perform dimension reduction while maintaining spatial topology and locality. Many researchers~\cite{Efficienthilbert,Ptv3,Octformer,Flatformer,DSVT,pointgpt,lest} have introduced space-filling curves for point cloud processing. HilbertNet~\cite{Efficienthilbert} uses the Hilbert curve to collapse 3D structures into 2D space to reduce computation and GPU occupation. 
% Wang \textit{et al.}~\cite{hilbert_point_index} utilize octree and hilbert curve to compress point clouds information. 
% C-Flow~\cite{C_flow} re-orders the 3D point clouds according to a Hilbert sorting scheme. 
PointGPT~\cite{pointgpt} utilizes the Morton-order curve~\cite{morton_order_curve} to introduce sequential properties.
OctFormer~\cite{Octformer} preserves Z-order during octreelization and adopts octree-attention for efficient context learning. PTV3~\cite{Ptv3} streamlines the complex interaction with the space-filling curve serialization. For 3D object detection, some methods~\cite{DSVT,Flatformer} employ window sweep curves to group voxel features for parallel computation. We employ the Hilbert curve due to its advantageous characteristic of locality preservation.

\textbf{Point Cloud Grouping.} LiDAR point clouds are sparsely and non-uniformly distributed with varying densities. 
% The point sequences significantly exceed that of images. 
Therefore, existing methods group points or voxels to facilitate parallel computation and reduce complexity.
% commonly group points or voxels for parallel computation and reducing computational complexity. 
In point cloud analysis, some works~\cite{Pointnetpp, dgcnn} use the $K$ nearest neighbor (KNN) method to create groups of query points. 
However, the heavy computation burden makes KNN hard to scale for outdoor scenes. 
For 3D object detection, VoTr~\cite{VoxelTransformer} uses a GPU-based hash table to search neighborhoods and generate fixed-length voxel groups. 
% Some methods~\cite{SST,swformer} group the point voxels within a window into different batches. 
Window-based Voxel Transformers~\cite{SST,swformer,DSVT,Flatformer} group voxels by employing a window-based sorting strategy, such as the rotating partition.
% More effectively, some works~\cite{DSVT,Flatformer} adopt a window-based sorting strategy for grouping point pillars from different axes. 
To reduce the reliance on relative position in grouping operations, some recent works~\cite{Octformer,Ptv3,lest} have been proposed to group voxels based on space-filling curves.
% while preserving certain spatial proximity. 
However, grouping is merely a compromise for computational complexity, which restricts the flow of information and effective receptive field. To tackle this problem, we model the entire voxels into one single sequence and allow each voxel be aware of global context information.

% I do this work later
\label{sec:relatedwork}

\section{Methods}

In this section, we present Voxel Mamba, a group-free Voxel State Space Model-based 3D backbone that can be applied to most voxel-based 3D detectors. We first introduce the preliminary concepts associated with our method, followed by the overall architecture of Voxel Mamba. Then, we describe in detail the fundamental components of Voxel Mamba, including the Hilbert Input Layer (HIL), Dual-scale SSM Block (DSB), and Implicit Window Partition (IWP). 
% Finally, we thoroughly analyzed the design of our UnSSMs block alongside other potential architectures.

\begin{figure*}[t!]
     \centering
    \includegraphics[width=1\textwidth]{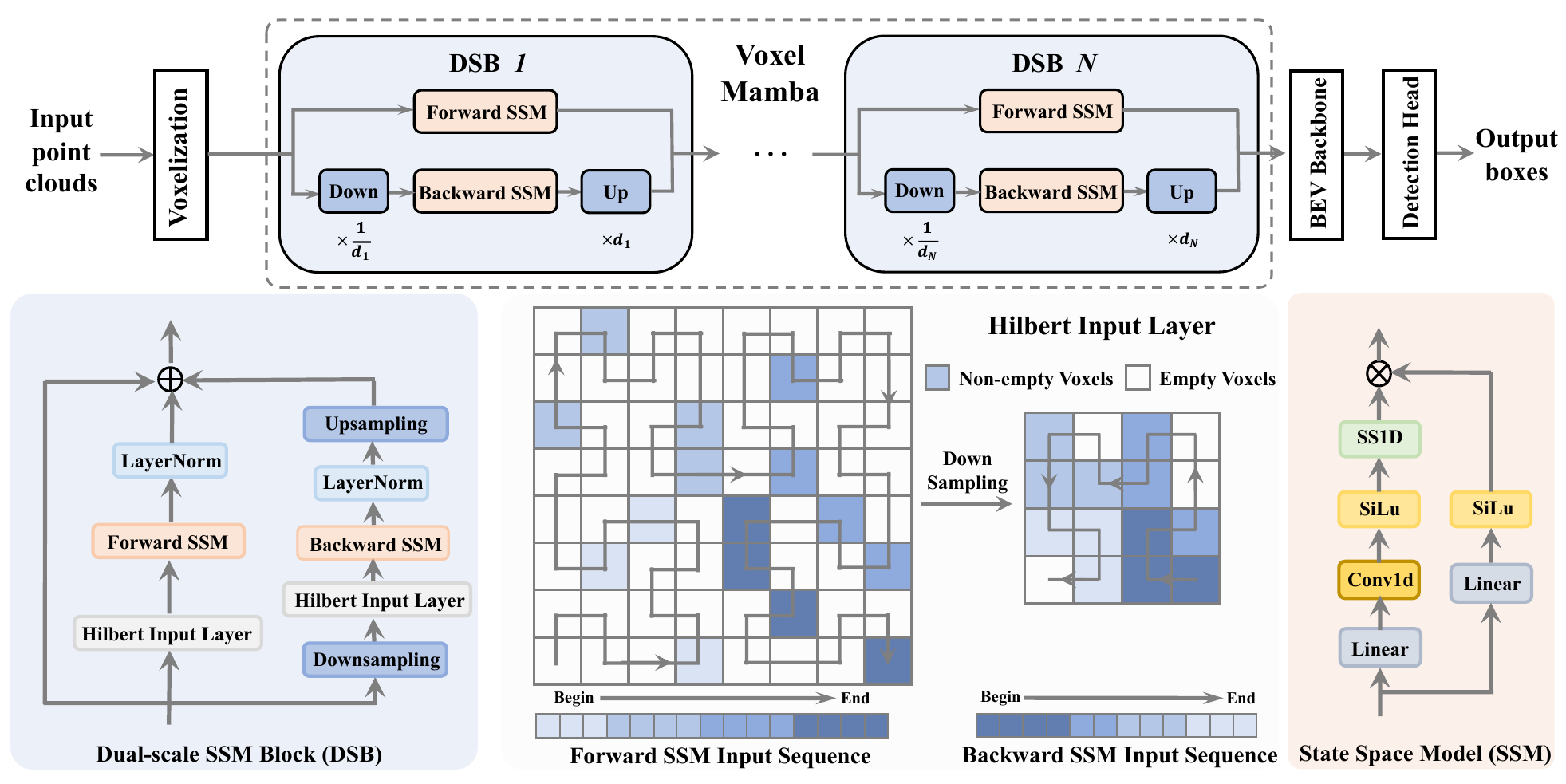}
    % \vspace{-1mm}
     \caption{\textbf{Top:} The overall architecture of our proposed Voxel Mamba with $N$ Dual-scale SSM Blocks (DSBs). \textbf{Bottom:} Illustration of the DSB, including a residual connection, a forward SSM branch, and a backward SSM branch.
     %
     % For an incoming set of tokens, Hilbert Input Layer converts them into a group-free sequence. Then, two SSM branches deal with voxel sequences in different resolution representation. 
     }
    \label{fig:main}
    % \vspace{-6mm}
\end{figure*}
% \vspace{3mm}
\subsection{Preliminaries}

The state space sequence (SSM) model is a continuous-time latent state model, which maps
a 1D input signal $x(t)\in \mathbb{R}^{L}$ to an output signal $y(t)\in\mathbb{R}^{L}$ through hidden state $h(t)\in \mathbb{R}^{N}$. The system can be represented as the following linear ordinary differential equation:
\begin{equation}
\label{eq::SSMs}
\left\{
\begin{aligned}
h'(t)=\mathbf{A}h(t)+\mathbf{B}x(t), \\
y(t)=\mathbf{C}h(t)+\mathbf{D}x(t),
\end{aligned}
\right.
\end{equation}
where $\mathbf{A} \in\mathbb{R}^{N\times N}$, $\mathbf{B} \in\mathbb{R}^{N\times 1}$ and $\mathbf{C} \in\mathbb{R}^{1\times N}$ are learnable parameters, and $\mathbf{D}\in\mathbb{R}^{1} $ denotes a residual connection.

To apply SSM to a discrete sequence, we can discrete the continuous-time SSM with a timescale parameter $\mathbf{\Delta}$ ~\cite{S4,Mamba,Vmamba}. The zero-order hold (ZOH) transformation can be used to discrete the continuous parameters $\mathbf{A, B}$ as $\mathbf{\overline{A}=\exp(\mathbf{\Delta A}),\overline{B}=(\mathbf{\Delta A})^{-1}(\exp(\mathbf{\Delta A}) - I) \cdot \mathbf{\Delta A}}$.
% \begin{equation}
% \label{eq::discrete AB}
% \begin{aligned}
% &\mathbf{\overline{A}} = \exp(\mathbf{\Delta A}) \\
% &\mathbf{\overline{B}} = (\mathbf{\Delta A})^{-1}(\exp(\mathbf{\Delta A}) - I) \cdot \mathbf{\Delta A}\\
% &\mathbf{\overline{C}}=\mathbf{C}, \mathbf{\overline{D}}=\mathbf{D}
% \end{aligned}
% \end{equation}
The discretized version of Eq.\eqref{eq::SSMs} can be written in the following recurrent form:
\begin{equation}
\label{eq::SSMs_discreted}
\left\{
\begin{aligned}
&h_{k}=\mathbf{\overline{A}}h_{k-1}+\mathbf{\overline{B}}x_{k}, \\
&y_{k}=\mathbf{\overline{C}}h_{k}+\mathbf{\overline{D}}x_{k}.
\end{aligned}
\right.
\end{equation}
Finally, the convolutional mode can be used for efficient parallel training:
\begin{equation}
\label{eq::SSMs_CNN}
\left\{
\begin{aligned}
&\mathbf{\overline{K}=(C\overline{B}, CA\overline{AB}, ..., C\overline{A}^{L}\overline{B}),} \\
&\mathbf{y = x \ast \overline{K},}
\end{aligned}
\right.
\end{equation}
where L is the length of the input sequence and $\overline{K}\in \mathbb{R}^{L}$ is the structured convolution kernel.

SSM combines the advantages of convolution and self-attention with near-linear computation and dynamic weights. It demonstrates stronger ability than Transformers in modeling long-range dependencies~\cite{S4,Mamba}, which inspires us to develop a group-free framework for point cloud based 3D object detection.

% Besides, the parameters in Eq.\ref{eq::SSMs_discreted} are fixed for all time steps, indicating linear time invariance (LTI). 
% Recent work~\cite{Mamba} introduces the selection mechanism into models to support dynamic weights. 
\subsection{Overall Architecture}

An overview of our proposed Voxel Mamba is shown in Figure~\ref{fig:main}. As in previous works~\cite{DSVT,Centerpoint,Pointpillar}, Voxel Mamba transforms point clouds into sparse voxels by a voxel feature encoding strategy. 
Unlike prior Transformer-based methods that perform extensive window partitioning and voxel grouping, in Voxel Mamba we serialize the voxel of the entire scene into a single sequence by using the \textit{Hilbert Input Layer} (Sec.~\ref{sec:hilbert}). 
Then, a \textit{Dual-scale SSM Block} (Sec.~\ref{sec:UnSSMs}) working on the voxel sequence is proposed, which allows voxels to be processed with a global context. 
% In contrast to previous Voxel Transformers ~\cite{SST,DSVT,Flatformer} that work on single-stride voxel features, ASSMs incorporate multi-scale contexts by working on the original voxel sequence in the forward path and down-sampled voxels sequence in the backward path. 
To enlarge the effective receptive fields, DSB adopts a finer-grained perception of the voxel sequence in the forward path, and down-samples the voxels sequence in the backward path. 
% Thus, the backward branch complements the global view and boosts performance with multi-scale semantic information. 
% It can complement the forward branch with varying resolution scene context. 
The backward path extracts features from the low-resolution BEV representation, with an increased downsampling factor in deeper blocks.
% adapts an increasing downsampling factor on bird's eye view (BEV) with the network depth.
% hierarchical architecture is utilized for the backward SSMs branch and built by downsampling the feature map on bird's eye view (BEV). 
% Inspired by previous window-partition methods, Voxel Mamba incorporates \textit{implicit window-based position embedding} (Sec.~\ref{sec:PIWP}) to capture richer position information and preserve the locality of voxel sequences.
To enhance the spatial proximity in sequences, Voxel Mamba adopts \textit{Implicit Window Partition} (Sec.~\ref{sec:PIWP}) to preserve 3D positional information in the extracted voxel features, and projects them to a BEV feature map.
% capture richer position information for voxel features in sequences. 
% into voxel tokens within each window-based region to enhance local context in long sequences (three orders of magnitude larger than typical grouping sequences).
Our proposed architecture is flexible and can be applied to most existing 3D object detection frameworks.
% Voxel Mamba has three stages, each of which contains two ASSM blocks. The feature maps in each stage are downsampled by factors of 1, 2, and 4, respectively. With this design, several UnSSMs blocks are stacked to apply on voxels.

% \begin{figure*}[t!]
%      \centering
%     \includegraphics[width=1\textwidth]{image/comparsion2.pdf}
%     % \vspace{-1mm}
%      \caption{Comparsion of different architecture.}
%     \label{fig:structure_insight}
%     % \vspace{-3mm}
% \end{figure*}
% \vspace{3mm}
\subsection{Hilbert Input Layer}
\label{sec:hilbert}

% Converting voxels from 3D space into a sequence inevitably reduces spatial proximity. 
The space-filling curve (\textit{e.g.,} Hilbert~\cite{hilbertcurve} and Z-order~\cite{zorder_curve}), known for preserving spatial locality, is widely used for dimensionality reduction.
% As shown in Figure~\ref{fig:main}, 
Space-filling curves, such as the Hilbert shown in Fig.~\ref{fig:main},  can traverse all elements in a space without repetition and preserve spatial topology. 
To improve the voxel proximity in serialization, we propose the \textit{Hilbert Input Layer} to reorder the voxel sequence. %Hilbert is selected due to its widespread use as a space-filling curve.

Denote the coordinates of voxel features as $\mathcal{C}=\{(x,y,z)\in\mathbb{R}^{3} | 0 \leq x,y,z \leq n\}$. We map a voxel onto its traversal position $h$ within the Hilbert curve. Specifically, we transform $(x,y,z)$ into its binary format with $log_2 n$ bits. For example, $x$ is converted to $(x_{m}x_{m-1}...x_{0})$, where $m=\lfloor log_2 n \rfloor$. Then, following~\cite{programming_hilbert}, we iterate from $x_m,y_m,z_m$ to $x_1,y_1,z_1$ bits and perform exchanges and inversions to adjust the order of bits. An exchange is conducted when the current bit is 0; otherwise, an invert is conducted. We concatenate all bits as $(x_{m}y_{m}z_{m}x_{m-1}y_{m-1}z_{m-1}\ldots x_{0}y_{0}z_{0})$ and apply a global 3$m$-fold Gray decoding~\cite{programming_hilbert} on it to obtain the traversal position $h$. Subsequently, all voxels are sorted into a single sequence based on their traversal position $h$.
% Thus, \textit{Hilbert Input Layer} can effectively map voxels into a group-free sequence while preserving the ordering of the Hilbert curve. 

In our implementation, we record the traversal position $h$ corresponding to the coordinates of all potential voxels.
The voxels are serialized by querying and sorting their traversal positions. 
Notably, the serialization process only takes approximately 0.7ms for a sequence of length $10^6$.

\subsection{Dual-scale SSM Block}
\label{sec:UnSSMs}
Though  space-filling curves can preserve the 3D structure to a certain degree, proximity loss is inevitable due to the dimension collapse from 3D to 1D.
As a result, a local snippet of the curve can only cover a partial region in 3D space.
As discussed in Sec.~\ref{sec:intro}, placing all voxels in a single group cannot ensure that the effective receptive field (ERF)~\cite{erf,replknet} could cover all voxels.
%So, without sufficient ERF, the network is likely to learn from the suboptimal local features.
Therefore, in this subsection we introduce the Dual-scale SSM block (DSB) to build a hierarchy of state space structures and consequently improve the ERF of the model.

% After enhancing proximity and locality in both features and sequence spaces, we aim to further optimize Voxel Mamba architectures to improve spatial coherence.
As shown in Fig.~\ref{fig:main}, the DSB block is designed with a residual connection ~\cite{ResNet}, a forward SSM branch and a backward SSM branch. It operates on two serialized voxel sequences generated by the Hilbert Input Layer, enabling a seamless flow of information throughout the voxel sequence. The forward branch processes the original voxel sequence, maintaining high-resolution details.
% and ensuring a comprehensive analysis of the scene. 
The backward branch, however, operates on a down-sampled voxel sequence derived from a low-resolution BEV representation. 
This dual-scale path allows DSB to incorporate larger-scale voxel features, enhancing the model's ability to model long dependencies among voxels. 
Specifically, given a voxel sequence $\mathcal{F}$ and its corresponding coordinates $\mathcal{C}$, DSB is computed as:
\begin{equation}
\begin{aligned}
    &\mathcal{F}_{f} = \mathbf{LN}(\mathbf{FSSM}(\mathbf{HIL}(\mathcal{F} + \mathbf{IWE}(\mathcal{C})))), \\
    &\mathcal{F}_{b} = \mathbf{Up}(\mathbf{LN}(\mathbf{BSSM}(\mathbf{HIL}(\mathbf{Down}(\mathcal{F}) + \mathbf{IWE}(\mathcal{C}^{'}) )))),\\
    &\widetilde{{\mathcal{F}}}=\mathcal{F}_{f} + \mathcal{F}_{b} + \mathcal{F},
\end{aligned}
\end{equation}
where $\mathbf{HIL}(\cdot)$ represents the Hilbert Input Layer, $\mathbf{FSSM}(\cdot)$ and $\mathbf{BSSM}(\cdot)$ denote the forward and backward SSM, $\mathbf{LN}(\cdot)$ stands for Layer Normalization, and $\mathcal{C}^{'}$ is the coordinates of downsampled sparse voxels. Besides, $\mathbf{Down}(\cdot)$ and $\mathbf{Up}(\cdot)$ refer to the downsampling and upsampling operations, respectively, and $\mathbf{IWE}(\cdot)$ means Implicit Window Embedding.
Overall, DSB integrates the widely adopted bidirectional design~\cite{Vision_Mamba,Vmamba} with the hierarchical design, building sufficient receptive field to mitigate the loss of proximity without introducing additional parameters.

\subsection{Implicit Window Partition}
\label{sec:PIWP}

% Although the Hilbert Input Layer, we serialize all the 3D voxels into a single sequence.
% However, as discussed in Sec.~\ref{sec:intro}, placing all voxels in a single group does not mean that the effective receptive field is able to cover all voxels.
The window partition strategy is widely used in previous 3D detectors~\cite{SST, DSVT} to enhance the voxel proximity.
In these methods, the whole field is partitioned into multiple local windows and the voxels within a window form a group. Therefore, the voxels inside a window will have sufficient proximity; however, the voxels in different windows will have minimal proximity.
In this section, we aim to introduce the advantages of window partition into our framework while avoiding its weaknesses. 

% can improve the voxel proximity in sequence space, the SSM input features still lack 3D local dependencies and position information.
To fulfill our goal, we propose an Implicit Window Partition (IWP) strategy. 
Unlike previous methods, we do not explicitly partition voxels into windows and apply Transformer or SSM within each window.
In contrast, we calculate the voxel coordinates inside and across windows, and then encode coordinates to embeddings, termed as Implicit Window Embedding (IWE), which is formulated as:
% Inspired by the success of position embedding in CNN~\cite{condinst,coordconv} and Transformers~\cite{swin_transformer,ViT}, we employ it to enhance the locality information in voxel features.
% Previous group-based detectors~\cite{SST,DSVT,Flatformer} demonstrate the efficacy of window partition in preserving the locality. However, these explicit window-based partitions necessitate extensive sorting and padding operations to prepare for further processing. We argue that window-based coordinates are essential for preserving local properties. Inspired by this, we do implicit window partition by position embedding to capture richer position information for voxel features as follows:
% However, we discover that only window partition-based coordinates can capture rich local dependencies. 
% Besides, position embedding has demonstrated its effectiveness in both CNN~\cite{condinst,coordconv} and Transformers~\cite{swin_transformer,ViT} frameworks.
% Previous grouping-based methods, such as SST~\cite{SST} and DSVT~\cite{DSVT}, have shown the efficacy of window partition in preserving the locality and geometry in 3D detectors. 
% Inspired by this, we do implicit window partition by position embedding to capture richer position information for voxel features as follows:
% window-based region position embedding to capture richer position information. 
\begin{equation}
\begin{aligned}
    &\mathbf{IWE} = \mathbf{MLP}(\text{concat}(z, \lfloor \frac{x^{i}}{w} \rfloor, \lfloor \frac{y^{i}}{h} \rfloor, x^{i}\ \text{mod}\ w, y^{i}\ \text{mod}\ h)), i=0,1 \\ 
    % \lfloor \frac{x_{j}, y_{j}}{w} \rfloor, (x_{j},y_{j})\ \text{mod}\ w))\\
    % \mathbf{LN}(\mathbf{SSM}(\mathbf{HIL}(\mathcal{F}))), \\
\end{aligned}
\end{equation}
where $\lfloor \cdot \rfloor$ is the floor function, $w,h$ define the window shape, and $z, x^{i}, y^{i}$ are the coordinates of tokens. $(x^{0}, y^{0})$ and $(x^{1}, y^{1})$ represent the coordinates before and after an implicit window shift. 
The IWE is shared across all layers with the same stride.
Thus, its computation cost only comes from shallow MLPs. With IWE, voxels in the serialized 1D curve are aware of their positions and consequently their proximity in 3D space.

\subsection{The Voxel Mamba Backbone}
% \textbf{Backbone.} 
With the proposed Hilbert Input Layer, DSB and IWP strategies, we build Voxel Mamba, a group-free sparse voxel backbone. 
The architecture of Voxel Mamba is illustrated in Figure~\ref{fig:main}. 
It comprises $N$ DSB blocks, which are organized into different stages based on their downsampling rates.
SpConv~\cite{spconv} is employed to progressively decrease the feature map resolution along the Z-axis in each stage.
Before sparse tokens are fed into the BEV backbone, we scatter them into dense BEV features.
On the Waymo dataset, we adopt the BEV backbone from Centerpoint-Pillar~\cite{Centerpoint}, and employ the same setting as DSVT~\cite{DSVT} for the detection head and loss function.
% In terms of detection head and loss function, we simply follow DSVT~\cite{DSVT}.
On the nuScenes dataset, we only replace the 3D backbone of DSVT~\cite{DSVT} with our Voxel Mamba backbone.

% To support effective 3D downsampling

% We develop both pillar-based and voxel-based models to work with the existing detector heads.
% SpConv~\cite{spconv} is employed to progressively decrease the feature map size along the Z-axis in each stage, and we scatter sparse tokens back to dense BEV feature maps at the last block.
% The output voxel number and corresponding coordinates of those two versions, pillar-based and voxel-based, are equal. 
% On the Waymo dataset, the Centerpoint-Pillar~\cite{Centerpoint} is adopted as the BEV feature extractor. In terms of detection head and loss function, we simply follow DSVT~\cite{DSVT}. On the nuScenes dataset, we only replace the 3D backbone of Transfusion-L~\cite{transfusion} with Voxel Mamba. 
% (How about Waymoo?)

% \subsection{Discussion}
% \label{sec:method:discussion}
% Many other architectures can incorporate multiscale features while ensuring resolution requirements. As shown in Figure~\ref{fig:structure_insight}, we enumerate some other potential structures, such as Bottleneck Bidirectional Mamba.

\label{sec:method}

\section{Experiments}

\begin{table*}[t]
    \caption{Performance comparison on the \textbf{validation} set of Waymo Open Dataset (single-frame setting). Symbol `-' means that the result is not available.}
    \centering
    \resizebox{1\columnwidth}{!}{
        \begin{tabular}{l|c|cccccccc}
        \toprule
         &  & \multicolumn{2}{c}{ALL (3D mAPH)} & \multicolumn{2}{c}{Vehicle (AP/APH)} & \multicolumn{2}{c}{Pedestrian (AP/APH)} & \multicolumn{2}{c}{Cyclist (AP/APH)} \\
        \multirow{-2}{*}{Method} & \multirow{-2}{*}{Category} & L1 & L2 & L1 & L2 & L1 & L2 & L1 & L2 \\
            \midrule
            PointPillar \cite{Pointpillar}& \multirow{3}{*}{2D CNN} &63.3& 57.5& 71.6 / 71.0& 63.1 / 62.5& 70.6 / 56.7& 62.9 / 50.2& 64.4 / 62.3& 61.9 / 59.9\\
            Centerpoint-Pillar \cite{Centerpoint} & & -& - & 76.1 / 75.5 & 68.0 / 67.5& -& - / 62.6& -& - / 67.6\\
            PillarNeXt \cite{pillarnext} & & 75.7& 69.7&78.4 / 77.9& 70.3 / 69.8&82.5 / 77.1& 74.9 / 69.8 &73.2 / 72.2&70.6 / 69.6 \\
            \midrule
            % IA-SSD \cite{ia-ssd} & 1& 64.48&58.08&70.53 / 69.67& 61.55 / 60.80 &69.38 / 58.47& 60.30 / 50.73& 67.67 / 65.30 &64.98 / 62.71\\
            % LiDAR R-CNN \cite{Lidarrcnn}& 66.2& 60.1& 73.5 / 73.0& 64.7 / 64.2& 71.2 / 58.7& 63.1 / 51.7& 68.6 / 66.9& 66.1 / 64.4\\
            % SpCNN
            SECOND \cite{Second} & \multirow{11}{*}{SpCNN} &63.1& 57.2 &72.3 / 71.7& 63.9 / 63.3& 68.7 / 58.2& 60.7 / 51.3& 60.6 / 59.3& 58.3 / 57.1 \\
            Part-A2 \cite{part-aware}& &70.3& 63.8& 77.1 / 76.5& 68.5 / 68.0& 75.2 / 66.9& 66.2 / 58.6& 68.6 / 67.4& 66.1 / 64.9\\
            PV-RCNN \cite{PVRCNN}&  &69.6& 63.3& 77.5 / 76.9& 69.0 / 68.4& 75.0 / 65.7& 66.0 / 57.6& 67.8 / 66.4& 65.4 / 64.0\\
            Centerpoint-Voxel \cite{Centerpoint} & & - & 67.6 & 76.6 / 76.0 & 68.9 / 68.4& 79.0 / 73.4 & 71.0 / 65.8 & 72.1 / 71.0 & 69.5 / 68.5\\
            PV-RCNN++ \cite{PVRCNN++} & & 75.2& 68.6& 79.1 / 78.6& 70.3 / 69.9& 80.6 / 74.6& 71.9 / 66.3& 73.5 / 72.4& 70.7 / 69.6\\
            AFDetV2\cite{Afdetv2} &  &74.8 & 68.8 & 77.6 / 77.1 & 69.7 / 69.2 & 80.2 / 74.6 & 72.2 / 67.0 & 73.7 / 72.7 & 71.0 / 70.1 \\
            VoxelNeXt \cite{VoxelNext} & & 76.3 & 70.1 & 78.2 / 77.7 & 69.9 / 69.4 &81.5 / 76.3 &73.5 / 68.6 &76.1 / 74.9 &73.3 / 72.2\\
            HEDNet~\cite{hednet} &  &79.4 & 73.4 & 81.1 / 80.6 & 73.2 / 72.7 & 84.4 / 80.0 & 76.8 / 72.6 & 78.7 / 77.7 & 75.8 / 74.9\\
            PillarNet \cite{pillarnet}& &74.6&68.4&79.1 / 78.6& 70.9 / 70.5& 80.6 / 74.0& 72.3 / 66.2& 72.3 / 71.2& 69.7 / 68.7\\
            FSD \cite{FSD} &  & 77.3 & 70.8 & 79.2 / 78.8 & 70.5 / 70.1 & 82.6 / 77.3 & 73.9 / 69.1 & 77.1 / 76.0 & 74.4 / 73.3 \\
            ConQueR~\cite{conquer} & & 77.9 & 71.6 & 78.4 / 77.9 & 71.0 / 70.5 & 82.4 / 76.6 & 75.8 / 70.1 & 77.5 / 76.4 & 75.2 / 74.1 \\
            % RSN \cite{rsn}& 1& -& -& 75.10 / 74.60& 66.00 / 65.50& 77.80 / 72.70& 68.30 / 63.70& -& -\\
            \midrule
            VoTR \cite{VoxelTransformer}& \multirow{9}{*}{Group-based} & -&-&75.0 / 74.3&65.9 / 65.3&-&-&-&-\\
            VoxSeT \cite{Voxset}& & 72.2 & 66.2 &74.5 / 74.0&66.0 / 65.6&80.0 / 72.4&72.5 / 65.4&71.6 / 70.3&69.0 / 67.7\\
            SST\cite{SST}& & -& -&76.2 / 75.8& 68.0 / 67.6&81.4 / 74.1& 72.8 / 65.9&-&-\\
            SWFormer\cite{swformer} & & -& -&77.8 / 77.3& 69.2 / 68.8&80.9 / 72.7 &72.5 / 64.9 &-&- \\
            CenterFormer~\cite{centerformer} & & 73.2 & 69.1 & 75.0 / 74.4 & 69.9 / 69.4 & 78.0 / 72.4 & 73.1 / 67.7 & 73.8 / 72.7 & 71.3 / 70.2 \\
            FlatFormer \cite{Flatformer} & & - & 67.2 & - & 69.0 / 68.6 & - & 71.5 / 65.3 & -  & 68.6 / 67.5 \\
            PTv3 \cite{Ptv3} & &- & 70.5 & -  & 71.2 / 70.8 & - & 76.3 / 70.4 & - & 71.5 / 70.4 \\
            % FocalFormer3D~\cite{focalformer3d} & - & 69.0 & - & - / 67.6 & - & - / 66.8 & - & - / 72.6 \\
            DSVT-Voxel \cite{DSVT}&  & 78.2 & 72.1 & 79.7 / 79.3& 71.4 / 71.0& 83.7 /     78.9& 76.1 / 71.5& {77.5} / 76.5& 74.6 / 73.7\\
            % Pillar original setting
            % \textbf{Voxel Mamba} (pillar)  & 1 & {78.07} & {71.11} & 79.78 / 79.33 & 71.70/ 71.28 & 83.42 / 77.78 & 76.05 / 70.65 & 75.19 / 74.16 & 72.38 / 71.39 \\
        % \textbf{Voxel Mamba} (voxel 20\%)  & 1 &\textbf{} & \textbf{} &            \textbf{79.18} / \textbf{78.71}& \textbf{70.77} / \textbf{70.34}&           \textbf{84.74} / \textbf{79.98}& \textbf{77.49} / \textbf{72.92}&           77.77 / \textbf{76.68}& \textbf{74.94} / \textbf{73.89}\\
            % 128 original setting
            % \textbf{Voxel Mamba}\textsuperscript{128} (voxel)  & 1 &\textbf{79.75} & \textbf{73.71} & {80.70} / {80.25}& {72.50} / {72.09}& {84.86} / {80.52}& {77.63} / {73.41}& \textbf{79.51} / \textbf{78.48}& \textbf{76.64} / \textbf{75.65}\\
            %%%%%% final results %%%%%%%
            % \textbf{Voxel Mamba} (voxel) & \textbf{79.57} & \textbf{73.35} & \textbf{80.78} / \textbf{80.34}& \textbf{72.62} / \textbf{72.21}& \textbf{85.00} / \textbf{80.79}& \textbf{77.72} / \textbf{73.64}& \textbf{78.56} / \textbf{77.59}& \textbf{75.74} / \textbf{74.80}\\
            \midrule
            \textbf{Voxel Mamba} (ours) & Group-free & \textbf{79.6} & \textbf{73.6} & {80.8} / {80.3}& {72.6} / {72.2}& {85.0} / {80.8}& {77.7} / {73.6}& {78.6} / {77.6}& {75.7} / {74.8}\\
            % 192 original setting
            % \textbf{Voxel Mamba}\textsuperscript{192} (voxel)  & 1 &{79.66} & {73.64} & \textbf{81.13} / \textbf{80.86}& \textbf{73.00} / \textbf{72.58}& \textbf{85.09} / \textbf{81.01}& \textbf{77.96} / \textbf{73.98}& {78.12} / {77.12}& {75.34} / {74.38}\\
            \bottomrule
        \end{tabular}}
	\label{tab:waymo_val}
		\vspace{-4mm}
\end{table*}

\subsection{Datasets and Evaluation Metrics}
\textbf{Waymo Open Dataset} contains 230k annotated samples, partitioned into 160k for training, 40k for validation and 30k for testing. Each frame covers a large perception range $(150m \times 150m)$. The  mean average precision (mAP) and its weighted variant by heading accuracy (mAPH) are used as evaluation metrics. They are further categorized into Level 1 for objects detected by over five points, and Level 2 for those detected with at least one point.

\textbf{nuScenes} consists of 40k labeled samples, with 28k for training, 6k for validation and 6k for testing. For 3D object detection, nuScenes employs the mean average precision (mAP) 
% across multiple distance thresholds (0.5, 1, 2, and 4 meters) 
and the nuScenes detection score (NDS) to measure model performance.

\begin{table*}[t]
		\caption{Performance comparison on the \textbf{test} set of Waymo Open Dataset. Symbol `-' means that the result is not available. "3f" stands for 3-frame model.}
  \vspace{-2mm}
		\centering
		\resizebox{1\columnwidth}{!}{
			\begin{tabular}{lcccccccc}
				 \toprule
     & \multicolumn{2}{c}{ALL (3D mAPH)} & \multicolumn{2}{c}{Vehicle (AP / APH)} & \multicolumn{2}{c}{Pedestrian (AP / APH)} & \multicolumn{2}{c}{Cyclist (AP / APH)} \\
            \multirow{-2}{*}{Method} & L1 & L2 & L1 & L2 & L1 & L2 & L1 & L2 \\
            \midrule
				PointPillar\cite{Pointpillar}&-&-& 68.6 / 68.1& 60.5 / 60.1& 68.0 / 55.5& 61.4 / 50.1&-&-\\
				% StarNet\cite{starnet}&-&-&61.0& 54.5& 67.8 / 59.9& 61.1 / 54.0 &-&-\\
				M3DETR\cite{m3detr} & 67.1 &61.9 &77.7 / 77.1& 70.5 / 70.0& 68.2 / 58.5& 60.6 / 52.0& 67.3 / 65.7& 65.3 / 63.8\\ 
				3D-MAN \cite{3dman} & -& -& 78.7 / 78.3& 70.4 / 70.0 & 70.0 / 66.0& 64.0 / 60.3& -& -\\
				PV-RCNN++ \cite{PVRCNN++} & 75.7& 70.2&81.6 / 81.2& 73.9 / 73.5& 80.4 / 75.0& 74.1 / 69.0& 71.9 / 70.8& 69.3 / 68.2 \\
				CenterPoint \cite{Centerpoint}& 77.2& 71.9& 81.1 / 80.6 &73.4 / 73.0& 80.5 / 77.3& 74.6 / 71.5& 74.6 / 73.7& 72.2 / 71.3 \\
				RSN \cite{RSN} & -& -& 80.7 / 80.3& 71.9 / 71.6& 78.9 / 75.6 &70.7 / 67.8 &-&-\\
				SST-3f \cite{SST} &78.3 &72.8 &81.0 / 80.6 &73.1 / 72.7& 83.3 / 79.7 &76.9 / 73.5 &{75.7} / {74.6} &{73.2} / {72.2} \\
                % HEDNet \cite{hednet} & 79.05 & 73.77 &83.78 / 83.39&76.33 / 75.96 &83.46 / 78.98	& 77.53 / 73.25& \textbf{75.86} / \textbf{74.78} & \textbf{73.13} / \textbf{72.09}\\
                Graph-RCNN~\cite{graphrcnn} & 77.0 & 71.6 & 83.6 / 83.1 & 76.0 / 75.6 & 81.9 / 76.5 & 75.6 / 70.5 & 72.5 / 71.3 & 69.8 / 68.7 \\ 
                FSDv1 \cite{FSD} & 78.2 & 72.4 & 82.7 / 82.3 & 74.4 / 74.1 & 82.9 / 77.9 & 75.9 / 71.3 & 75.6 / 74.4 & 72.9 / 71.8 \\
                FSDv2~\cite{fsdv2} & 79.0 & 73.3 & 82.4 / 82.0 & 74.4 / 74.0 & 83.8 / 78.9 & 77.4 / 72.8 & 77.1 / 76.0 & 74.3 / 73.2\\
                PillarNeXt-3f \cite{pillarnext} &79.0 & 74.1 & 83.3 / 82.8 & 76.2 / 75.8 & 84.4 / 81.4	& 78.8 / 76.0 & 73.8 / 72.7	& 71.6 / 70.6\\
				% \toprule[1pt]
                \textbf{Voxel Mamba} (ours) &\textbf{79.6}&\textbf{74.3}& {84.4} / {84.0} &{77.0} / {76.6} & {84.8} / {80.6} &{79.0} / {74.9}&{75.4} / {74.3}& {72.6} / {71.5}\\
				% \toprule[1pt]
                \bottomrule
			\end{tabular}}
            \vspace{-2mm}
		\label{tab:waymo_test}
		
\end{table*}

\subsection{Implementation Details}
Our method is implemented based on the open-source framework OpenPCDet~\cite{openpcdet}. 
The voxel sizes are defined as $(0.32m, 0.32m, 0.1875m)$ for Waymo and $(0.3m, 0.3m, 0.2m)$ for nuScenes.
We stack six DSB blocks, divided into three stages, for the Voxel Mamba backbone network.
%The state space model in Voxel Mamba follows standard SSM in Mamba~\cite{Mamba}.
The downsampling rates for DSBs' backward branches in each stage are $\{1,2,4\}$.
Specifically, we employ SpConv~\cite{spconv} and its counterpart SpInverseConv as downsampling and upsampling operators in the DSB backward branch. 
% For the voxel-based version, the architecture is similar to the pillar-based framework except for a finer division along the Z-axis. 
% The voxel sizes are defined as $(0.32m, 0.32m, 0.1875m)$ for Waymo and $(0.3m, 0.3m, 0.2m)$ for nuScenes. 
% We apply SpConv to downsample the voxel features exclusively along the Z-axis among the three stages.
On the Waymo dataset, Voxel Mamba is trained for $24$ epochs with a learning rate of $0.0025$.
% The voxel features have 128 channels.
On the nuScenes dataset, we train Voxel Mamba for $20$ epochs with a learning rate of $0.004$.
The voxel features on both datasets consist of 128 channels.
All the models are trained by the AdamW optimizer on 8 RTX A6000 GPUs. Other settings in the training and inference of Voxel Mamba follow DSVT~\cite{DSVT}.

\subsection{Comparison with State-of-the-art Methods}
% \begin{wrapfigure}{r}{0.5\textwidth}
% \vspace{-4mm}
%     \centering
%     \includegraphics[width=0.48\textwidth]{image/speed_performance_2.png}
%     \caption{Detection performance (mAPH/L2) vs. speed (FPS) on Waymo.}
%     \vspace{-3mm}
%     \label{fig:speed}
% \end{wrapfigure}

\textbf{Waymo.}
We first compare Voxel Mamba with state-of-the-art methods on the Waymo Open dataset.
Table~\ref{tab:waymo_val} shows the results on the validation set. Our proposed Voxel Mamba achieves 79.6/73.4 on L1/L2 mAPH, which are +1.4 and +1.5 better than DSVT-Voxel. Since our framework differs from DSVT only on the 3D backbone, it can be concluded that Voxel Mamba has superior ability in capturing voxel features.
In comparison with window-based (\textit{e.g.,} DSVT) or curve-based (\textit{e.g.,} PTv3) grouping methods, our group-free method Voxel Mamba consistently delivers better results.
% Compared with group-based, Voxel Mamba achieves the best performance on both L1/L2 mAPH.
Table~\ref{tab:waymo_test} shows the results on the test split. Voxel Mamba reaches 79.6/74.3 in terms of L1/L2 mAPH, which is even better than the 3-frame setting of PillarNeXt and SST.

% in Tab.~\ref{tab:waymo_val} and Tab.~\ref{tab:waymo_test}. 

\textbf{nuScenes.}
We then compare Voxel Mamba with previous state-of-the-art methods on nuScenes. Table~\ref{tab:nusc_val} shows the results on the validation set. Voxel Mamba achieves impressive results with 71.9 NDS and 67.5 mAP, which is +0.5 and +0.8 higher than the previous best method.
% Notably, due to the stronger BEV backbone in HEDNet, it is not a fair comparison with our approach.
Compared with DSVT, Voxel Mamba achieves +1.1 higher performance on mAP. The results on the test split are shown in Table~\ref{tab:nusc_test}. Our method also exhibits the best mAP and NDS.  

% \begin{wraptable}{r}{0.45\textwidth}
% \vspace{-4mm}
% \caption{Comparison with other well-designed architectures on GPU memory.}
%         \setlength{\extrarowheight}{3pt} % Increase the height of each row
% 		\resizebox{0.45\columnwidth}{!}{
%     \begin{tabular}{l|c|c}
% 		\hline
% 	    \multicolumn{1}{l}{Abalation} & \multicolumn{1}{c}{Backbone}  & Memory (GB)\\ 
%             \hline
%          PointPillar~\cite{Pointpillar} & \multirow{2}{*}{2D CNN} & 3.6  \\
%          Centerpoint-Pillar~\cite{Centerpoint} &  & 3.2\\
%          \hline
%          % Centerpoint-Voxel~\cite{Centerpoint} & \multirow{2}{*}{SpCNN} & 2.4\\
%          Part-A2~\cite{part-aware} & \multirow{2}{*}{SpCNN} & 2.9\\
%          % VoxelNeXt~\cite{PVRCNN} & &   \\
%          PV-RCNN++ (ResNet)~\cite{PVRCNN++} &    & 17.2\\ 
%          % DSVT-Pillar          & 3.8  & 64 & 79.5 / 77.1 & 73.2 / 71.0\\
%          \hline
%          SST~\cite{SST} & \multirow{2}{*}{Transformers} & 6.8\\
%          DSVT-Voxel~\cite{DSVT} &  & 4.2\\
%          % Voxel Mamba (Pillar) & 3.5  & 66 & 77.1 & 71.2\\
%          \hline
%          Voxel Mamba (Ours) & SSMs & 3.7  \\
%         % Other methods \\
% 	\hline
% 	\end{tabular}
%     }
% 		\label{tab:memory}
% \end{wraptable}
\textbf{Inference Efficiency.} We compare Voxel Mamba with other state-of-the-art methods in inference speed and performance accuracy in Fig.~\ref{fig:speed}. Notably, Voxel Mamba outperforms DSVT~\cite{DSVT} and PV-RCNN++~\cite{PVRCNN} by at least +1.5 in detection accuracy, while achieving faster speed. Some methods, such as CenterPoint~\cite{Centerpoint} and PointPillar~\cite{Pointpillar}, are faster than Voxel Mamba; however, their accuracy is substantially lower. 

We further compare Voxel Mamba with previous well-designed architectures (SpCNN, Transformers, and 2D CNN) in GPU memory in Table~\ref{tab:memory}. Compared with CenterPoint-Pillar, Voxel Mamba requires only an additional 0.5 GB GPU memory but achieves +9.0 higher accuracy in L2 mAPH.
While Transformer-based methods like SST~\cite{SST} and DSVT~\cite{DSVT} use group partitioning, they still consume more memory than our group-free Voxel Mamba.
All the experiments are evaluated on an NVIDIA A100 GPU with the same environment.

%% Nuscenes Results
\begin{table*}[t]
    \centering
    \caption{Comparison with the state-of-the-art detectors on the nuScenes dataset \textbf{validation} split. 
    % ‘T.L.’, ‘C.V.’, ‘Ped.’, ‘M.T.’, ‘T.C.’, and 'B.R.' denote trailer, construction vehicle, pedestrian, motor, traffic cone, and barrier, respectively.
    }
 % \vspace{-2mm}
    \setlength{\tabcolsep}{1.4mm}{}
% \resizebox{1\columnwidth}{!}
    \scalebox{0.9}{
        \begin{tabular}{lcccccccccccc}
        \toprule
        	% \hline\rowcolor{gray!20}
        Method & NDS & mAP & Car & Truck & Bus & T.L. & C.V. & Ped. & M.T. & Bike & T.C. & B.R. \\ 
        \midrule
        CenterPoint~\cite{Centerpoint} & 66.5 & 59.2 & 84.9 & 57.4 & 70.7 & 38.1 & 16.9 & 85.1 & 59.0 & 42.0 & 69.8 & 68.3 \\
        VoxelNeXt~\cite{VoxelNext} & 66.7 & 60.5 & 83.9 & 55.5 & 70.5 & 38.1 & 21.1 & 84.6 & 62.8 & 50.0 & 69.4 & 69.4 \\
        TransFusion-L~\cite{transfusion} & 70.1 & 65.5 & 86.9 & 60.8 & 73.1 & 43.4 & 25.2 & 87.5 & 72.9 & 57.3 & 77.2 & 70.3 \\
        % FSDv2~\cite{FSD} & 70.4 & 64.7 & 83.7 & 51.6 & 66.4 & 59.1 & 32.5 & 87.1 & 71.4 & 51.7 & 80.3 & 78.7 \\
        PillarNeXt~\cite{pillarnext} & 68.4 & 62.2 & 85.0 & 57.4 & 67.6 & 35.6 & 20.6 & 86.8 & 68.6 & 53.1 & {77.3} & 69.7 \\
        % FSDv2~\cite{fsdv2} & 70.4 & 
        HEDNet~\cite{hednet} & 71.4 & 66.7 & 87.7 & 60.6 & {77.8} & {50.7} & {28.9} & 87.1 & 74.3 & 56.8 & 76.3 & 66.9 \\
        DSVT~\cite{DSVT} & {71.1} & 66.4 & {87.4} & {62.6} & 75.9 & 42.1 & 25.3 & {88.2} & 74.8 & {58.7} & 77.8 & {70.9} \\
        % \hline
        % \textbf{Voxel Mamba} (pillar) & 70.5 & \textbf{66.9} & 87.1 & 60.4 & \textbf{77.8} & \textbf{48.7} & \textbf{28.9} & 87.7 & \textbf{76.2} & \textbf{59.8} & 76.1 & 66.5 \\ 
        
        % \textbf{Voxel Mamba} (pillar) & 71.1 & {65.8} & \textbf{88.0} & \textbf{62.8} & {75.6} & {40.9} & \textbf{26.3} & 87.8 & {75.0} & {57.4} & 77.4 & 66.7 \\ 
        
        % \textbf{Voxel Mamba} (voxel) & \textbf{71.8} & \textbf{67.0} & \textbf{88.0} & 62.7 & \textbf{76.6} & \textbf{44.3} & 25.5 & \textbf{89.2} & \textbf{75.8} & \textbf{58.8} & \textbf{80.9} & 68.4\\ 
        \textbf{Voxel Mamba} (ours) & \textbf{71.9} & \textbf{67.5} & {87.9} & {62.8} & {76.8} & {45.9} & 24.9 & {89.3} & {77.1} & {58.6} & {80.1} & {71.5}\\ 
        % \hline
        % Baseline & {71.3} & 66.7 & {87.8} & {61.8} & 77.1 & 44.3 & 27.4 & {89.00} & 76.3 & 56.6 & 79.9 & \textbf{67.8} \\
        % Pos Embed (linear)1 & {72.1} & 67.8 & {87.9} & {64.7} & 77.3 & 44.3 & 26.0 & {89.0} & 76.7 & 59.6 & 79.4 & \textbf{73.4} \\
        % Pos Embed (linear)2 & {71.8} & 67.6 & {87.6} & {62.7} & 75.4 & 46.3 & 26.3 & {88.5} & 76.8 & 61.1 & 79.3 & \textbf{71.8} \\
        \bottomrule
        \end{tabular}}
    \label{tab:nusc_val}
\vspace{-2mm}
\end{table*}

\begin{table*}[t]
\caption{Comparison with the state-of-the-art detectors on the nuScenes dataset \textbf{test} split.}
  \begin{center}
  \scalebox{0.9}{
     \setlength{\tabcolsep}{1.4mm}{}
     % \hspace{-4mm}
     \begin{tabular}{lcccccccccccc}
     \toprule
    Method & NDS & mAP & Car & Truck & Bus & T.L. & C.V. & Ped. & M.T. & Bike & T.C. & B.R.\\
    % \hdashline
    \midrule
    PointPillars~\cite{Pointpillar} & 45.3 & 30.5 & 68.4 & 23.0 & 28.2 & 23.4 & 4.1 & 59.7 & 27.4 & 1.1 & 30.8 & 38.9\\
    3DSSD~\cite{3dssd} & 56.4 & 42.6 & 81.2 & 47.2 & 61.4 & 30.5 & 12.6 & 70.2 & 36.0 & 8.6 & 31.1 & 47.9\\
    CenterPoint~\cite{Centerpoint} & 65.5 & 58.0 & 84.6 & 51.0 & 60.2 & 53.2 & 17.5 & 83.4 & 53.7 & 28.7 & 76.7 & 70.9\\
    FCOS-LiDAR~\cite{FCOS_LiDAR} & 65.7 & 60.2 & 82.2 & 47.7 & 52.9 & 48.8 & 28.8 & 84.5 & 68.0 & 39.0 & 79.2 & 70.7 \\
    AFDetV2~\cite{Afdetv2} & 68.5 & 62.4 & 86.3 & 54.2 & 62.5 & 58.9 & 26.7 & 85.8 & 63.8 & 34.3 & 80.1 & 71.0 \\
    UVTR-L~\cite{UVTR} & 69.7 & 63.9 & 86.3 & 52.2 & 62.8 & 59.7 & 33.7 & 84.5 & 68.8 & 41.1 & 74.7 & 74.9\\
    VISTA~\cite{vista} & 69.8 & 63.0 & 84.4 & 55.1 & 63.7 & 54.2 & 25.1 & 82.8 & 70.0 & 45.4 & 78.5 & 71.4\\
    Focals Conv~\cite{focalconv} & 70.0 & 63.8 & 86.7 & 56.3 & 67.7 & 59.5 & 23.8 & 87.5 & 64.5 & 36.3 & 81.4 & 74.1\\
    VoxelNeXt~\cite{VoxelNext} & 70.0 & 64.5 & 84.6 & 53.0 & 64.7 & 55.8 & 28.7 & 85.8 & 73.2 & 45.7 & 79.0 & 74.6 \\
    TransFusion-L~\cite{transfusion} & 70.2 & 65.5 & 86.2 & 56.7 & 66.3 & 58.8 & 28.2 & 86.1 & 68.3 & 44.2 & 82.0 & 78.2\\
    LinK~\cite{Link} & 71.0 & 66.3 & 86.1 & 55.7 & 65.7 & 62.1 & 30.9 & 85.8 & 73.5 & 47.5 & 80.4 & 75.5\\
    HEDNet~\cite{hednet} & {72.0} & {67.7} &{87.1} & 56.5 & {70.4} & {63.5} & {33.6} & 87.9 & 70.4 & 44.8 & 85.1 & {78.1} \\
    LargeKernel3D~\cite{largekernel3d} & 70.6 & 65.4 & 85.5 & 53.8 & 64.4 & 59.5 & 29.7 & 85.9 & 72.7 & 46.8 & 79.9 & 75.5\\
    PillarNet~\cite{pillarnet} & 71.4 & 66.0 &  {87.6} & 57.5 & 63.6 & 63.1 & 27.9 & 87.3 & 70.1 & 42.3 & 83.3 & 77.2 \\
    FSDv2~\cite{fsdv2} & 71.7 & 66.2 & 83.7 & 51.6 & 66.4 & 59.1 & 32.5 & 87.1 & 71.4 & 51.7 & 80.3 & 78.7 \\
    DSVT~\cite{DSVT} & 72.7 & 68.4 & 86.8 & {58.4} & 67.3 & 63.1 & {37.1} & 88.0 & 73.0 & 47.2 & 84.9 & {78.4}\\
    % \textbf{Voxel Mamba (voxel)} & \textbf{73.3} & \textbf{68.9} & 86.7 & {57.1} & {66.9} & {63.1} & {36.4} & \textbf{89.4} & \textbf{75.1} & \textbf{49.1} & \textbf{87.7} & {77.9} \\
    
    \textbf{Voxel Mamba} (ours) & \textbf{73.0} & \textbf{69.0} & 86.8 & {57.1} & {68.0} & {63.2} & {35.4} & {89.5} & {74.7} & {50.8} & {86.9} & {77.3} \\
    
     \bottomrule
     % \thickhline
     \end{tabular}}
  \end{center}
  \vspace*{-2mm}
  \label{tab:nusc_test}
\end{table*}

\begin{table*}[t]
  \centering
  \begin{minipage}{0.48\linewidth}
     \includegraphics[width=\textwidth]{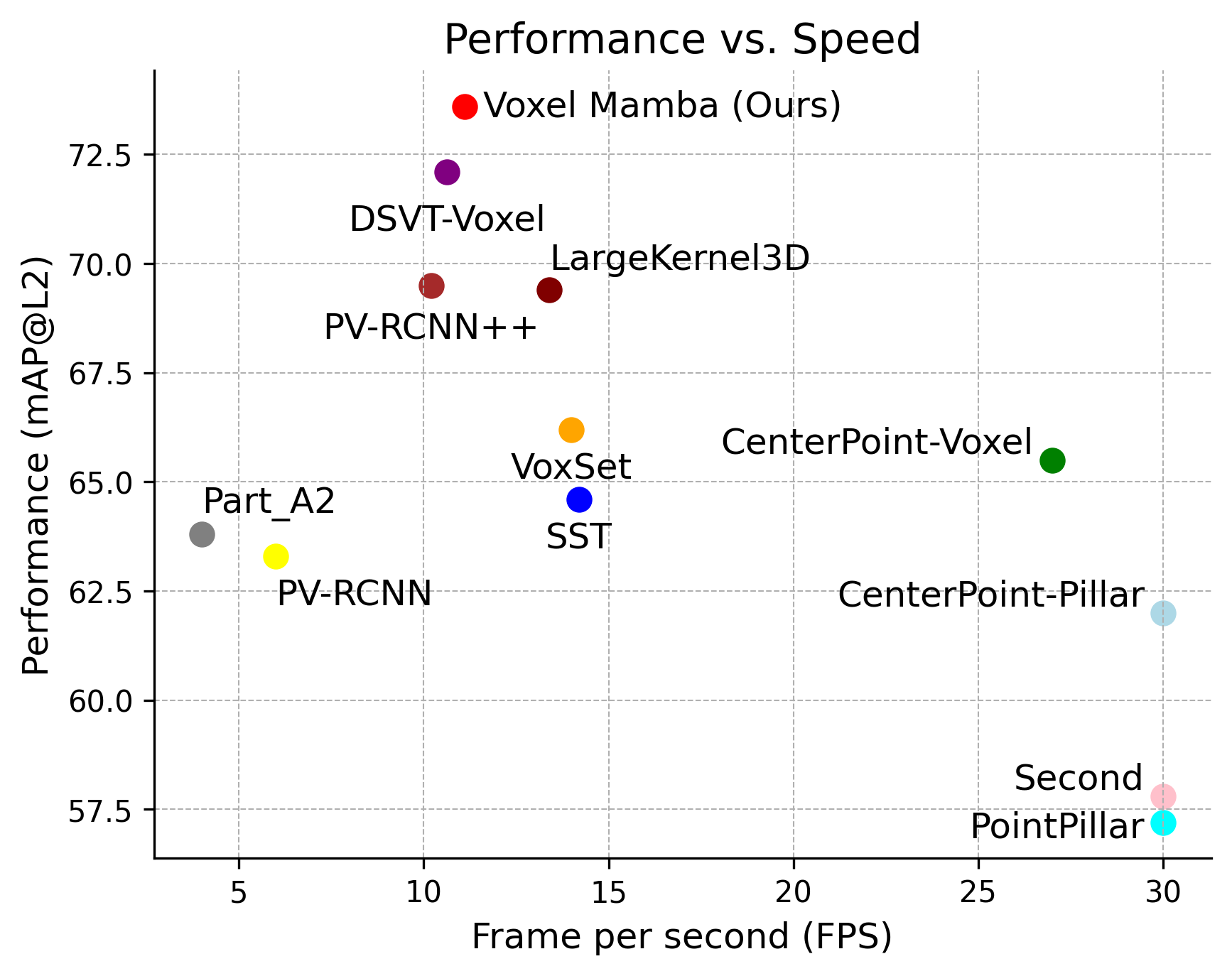}
     % \caption*{(a) Similarity.}
     % \vspace*{2mm}
     \refstepcounter{figure}\label{fig:speed}
  \end{minipage}\quad
  \begin{minipage}{0.48\linewidth}
    \centering
    \refstepcounter{table}\label{tab:memory}
    \setlength{\tabcolsep}{1mm}
    \setlength{\extrarowheight}{3pt}
    \resizebox{1\columnwidth}{!}{
    \begin{tabular}{l|c|c}
		\hline
	    \multicolumn{1}{l}{Abalation} & \multicolumn{1}{c}{Backbone}  & Memory (GB)\\ 
            \hline
         PointPillar~\cite{Pointpillar} & \multirow{2}{*}{2D CNN} & 3.6  \\
         Centerpoint-Pillar~\cite{Centerpoint} &  & 3.2\\
         \hline
         % Centerpoint-Voxel~\cite{Centerpoint} & \multirow{2}{*}{SpCNN} & 2.4\\
         Part-A2~\cite{part-aware} & \multirow{2}{*}{SpCNN} & 2.9\\
         % VoxelNeXt~\cite{PVRCNN} & &   \\
         PV-RCNN++ (ResNet)~\cite{PVRCNN++} &    & 17.2\\ 
         % DSVT-Pillar          & 3.8  & 64 & 79.5 / 77.1 & 73.2 / 71.0\\
         \hline
         SST~\cite{SST} & \multirow{2}{*}{Transformers} & 6.8\\
         DSVT-Voxel~\cite{DSVT} &  & 4.2\\
         % Voxel Mamba (Pillar) & 3.5  & 66 & 77.1 & 71.2\\
         \hline
         Voxel Mamba (Ours) & SSMs & 3.7  \\
        % Other methods \\
	\hline
	\end{tabular}
    }
  \end{minipage}
  \begin{minipage}{0.48\linewidth}
  \vspace{1mm}
  {Figure 3: Detection performance (mAPH/L2) vs. speed (FPS) on Waymo.}
  \end{minipage}
  \quad
  \begin{minipage}{0.48\linewidth}
  \vspace{1mm}
  { Table 5: Comparison with other well-designed architectures on GPU memory.}
  % \addtocounter{table}{1}
  \end{minipage}
\end{table*}

\begin{table*}[t]
  \caption{Ablations on the nuScenes validation split. In (d), Centerpoint-Pillar is used as the baseline.}
  \centering
  \begin{minipage}[t]{0.48\linewidth}
     \centering
     \setlength{\tabcolsep}{1mm}
     \resizebox{0.78\columnwidth}{!}{
     \begin{tabular}{lcc}
     \toprule
      Space-filing Curve  & mAP & NDS \\
        \midrule
        Random Curve & 66.0  &  71.0 \\
        Window Partition & 67.3 &  71.7\\
        Z-Order Curve & 67.0  &  71.7 \\
        Hilbert Curve & \textbf{67.5}  &  \textbf{71.9} \\
        \bottomrule
     \end{tabular}}
     % \vspace*{2mm}
  \end{minipage}\quad
  \begin{minipage}[t]{0.48\linewidth}
    \centering
    \setlength{\tabcolsep}{1mm}
    \resizebox{0.71\columnwidth}{!}{
     \begin{tabular}{lcc}
     \toprule
     Ablation & mAP & NDS  \\
     \midrule
        Baseline                    & 63.3 & 69.1  \\
        % \quad + bidirectional SSMs  & 65.7 & 70.8 \\
        % \quad + Hilbert curve       & 65.8 & 70.9 \\
        \quad + bidirectional SSMs  & 65.8 & 70.9\\
        \qquad  (Hilbert curve)      &\\
        \quad + Voxel               & 66.3 & 71.0  \\
        \quad + DSB                 & 66.7 & 71.3 \\
        \quad + IWP  & \textbf{67.5} & \textbf{71.9}\\
     \bottomrule
     \end{tabular}
     }
     % \vspace*{2mm}
  \end{minipage}
\begin{minipage}{0.48\linewidth}
  \centering
  \vspace{1mm}
  { \small (a) Ablation on space-filling curves.}
  \end{minipage}
  \begin{minipage}{0.48\linewidth}
  \centering
  \vspace{1mm}
  { \small (b) Effect of each component in Voxel Mamba.}
  \end{minipage}
  \begin{minipage}[t]{0.48\linewidth}
  \vspace{1mm}
    \centering
    \setlength{\tabcolsep}{0.5mm}
    \resizebox{0.56\columnwidth}{!}{
    \begin{tabular}{lcc}
      \toprule
      Downstrides  & mAP & NDS  \\ 
      \midrule
        \{1,1,1\} & 66.6  &  71.4 \\
        \{1,2,2\} & 66.9 & 71.8\\
        \{2,2,2\} & 65.6 & 70.8\\
        \{4,4,4\} & 66.2 & 71.2\\
        \{1,2,4\} & \textbf{67.5} &  \textbf{71.9}\\
      \bottomrule
     \end{tabular}
     }
  \end{minipage}\quad
  \begin{minipage}[t]{0.48\linewidth}
    % \hspace*{-6mm}
    \vspace{1mm}
     \centering
     \setlength{\tabcolsep}{1mm}
     \resizebox{0.72\columnwidth}{!}{
    \begin{tabular}{lcc}
      \toprule
      Pos Embeding & mAP & NDS  \\ 
      \midrule
      Baseline & 66.7 &  71.3\\
      Absolute position & 66.9  &  71.2 \\
      % Cos, Sin(without z) & 67.3 & 72.0\\
      Cos, Sin & 66.6 & 71.4\\
      % Window partition & 67.6 & 71.8\\
      Ours (w/o shift) & 67.3 & 71.9\\
      Ours & \textbf{67.5} & \textbf{71.9}\\
      \bottomrule
     \end{tabular}
     }
  \end{minipage}
  \begin{minipage}{0.48\linewidth}
  \centering
  \vspace{1mm}
  { \small (c) Ablation on the downsampling rates of DSB.}
  \end{minipage}
  \begin{minipage}{0.48\linewidth}
  \centering
  \vspace{1mm}
  { \small (d) Ablation on IWE.}
  \end{minipage}
  \label{tab:ablations}
  \vspace*{-1mm}
\end{table*}

% \begin{table*}[t]
%   \centering
%   \begin{minipage}{0.38\linewidth}
%     \vspace{7mm}
%      \includegraphics[width=\textwidth]{image/speed_performance.png}
%      % \includegraphics[width=\textwidth]{image/speed.pdf}
%      % \phantomcaption
%      \label{fig:seed}
%   \end{minipage}\quad
%   \begin{minipage}{0.58\linewidth}
%   \includegraphics[width=\textwidth]{image/ERF985.pdf}
%   \end{minipage}
%   \begin{minipage}{0.38\linewidth}
%   \centering
%   \vspace{-5mm}
%   Figure 3: Detection performance (mAPH/L2) vs speed (FPS).
%   \end{minipage}
%   \begin{minipage}{0.58\linewidth}
%   \centering
%   \vspace{-5mm}
%   Figure 4: The \textit{Effective Receptive Field (ERF)} of Voxel Mamba (left) and DSVT (right). 
%   \end{minipage}
%   % \caption{(a) Comparison of similarity matrix for 3D backbone features with different voxel size: Left: (0.64, 0.64, 6), right:(0.32, 0.32, 6). (b) Results of a pilot study on Waymo val split with 20\% data.}
%   \label{ablations}
%   \vspace*{-1mm}
% \end{table*}

\subsection{Ablation Studies}
\label{sec:ablation}
To better investigate the effectiveness of Voxel Mamba, we conduct a set of ablation studies by using the nuScenes validation set. We follow OpenPCDet~\cite{openpcdet} to train all models for 20 epochs.

\textbf{Effectiveness of Space-filling Curves.} 
There are some potential alternatives to Hilbert curve for preserving locality. Here, we compare Hilbert curve with some commonly used space-filling curves (Z-order~\cite{Octformer} and window partition~\cite{SST}) in 3D detection. 
% The window sweep follows the grouping method in \cite{SST}. 
As shown in Table~\ref{tab:ablations}(a), 
without using space-filling curves (\textit{i.e.}, the row of `Random Curve'), there will be a notable decline in performance, which indicates that spatial proximity is crucial in the group-free setting.
By using the Z-order curve and window partition to introduce spatial proximity, the mAP and NDS are much improved. The serialization based on the Hilbert curve can further enhance the model performance.  
% However, we have observed that Voxel Mamba with different spatial filling curves shows a similar performance. 
%However, We have also observed that the choice of curve has only a minor impact on performance.
%It suggests the Voxel Mamba's superior capability to model long dependencies, due to its stable performance across various curves.

% Besides, sequences without any space-filling curve guidance will degrade the performance.
% We compared the models built based on different space-filling curves in Tab.~\ref{tab:ablations}. 

\textbf{Effectiveness of Each Component.}
To more clearly illustrate the effectiveness of the different components in Voxel Mamba, we conduct experiments by adding each of them to a baseline, which is set to Centerpoint-Pillar~\cite{Centerpoint}. 
As shown in Table~\ref{tab:ablations}(b), bidirectional SSMs with a Hilbert-based group-free sequence can significantly improve the accuracy over the baseline, which validates the feasibility of our group-free strategy. Besides, converting pillar to voxel can enhance much the detector's performance without group size constraints. 
Voxel Mamba with DSB obtain better performance than the plain bidirectional SSMs. This is because DSB can build larger ERFs and mitigate the loss of proximity. 
Furthermore, IWE further boosts Voxel Mamba's performance for its capability in capturing 3D position information and increasing voxel proximity.
% It demonstrates the feasibility of SSM in tackling 3D detection tasks. 
% Compared with window sweep, Hilbert shows a greater capacity for preserving the locality. Furthermore, finer grid partitions on point clouds substantially improve detector performance. 
% This indicates that Voxel Mamba possesses a strong capability to model long dependencies. Finally, despite its high performance, our ASSMs and IWPE can further enhance Voxel Mamba's capabilities.

\textbf{Downsampling Rates of DSB.}
We evaluate the impact of different downsampling rates in DSB by adjusting the stride $\{d_{1},d_{2},d_{3}\}$ in the backward SSM branch at each stage. $d_{i}=1$ means the original resolution is used. The results are shown in Table~\ref{tab:ablations}(c). We see that transitioning from \{1,1,1\} to \{1,2,2\} and to \{1,2,4\} enhances performance due to an enlarged effective receptive field and improved proximity by using larger downsampling rates at late stages.
However, DSBs with \{2,2,2\} or \{4,4,4\} compromise performance compared to \{1,1,1\}, indicating that using larger downsampling rates at early stages will lose some fine details.
Thus, we set the stride as \{1,2,4\} to strike a balance between effective receptive fields and detail preservation.

% \textcolor{red}{Here, we investigate the effect of downsampling factors in \textit{Dual-scale SSMs Block} in Table~\ref{tab:ablations}(c). First, we respectively define the down strides of the backward SSM branch in each stage as $\{d_{1},d_{2},d_{3}\}$. $d_{i}=1$ denotes that voxel features keep the original resolution. Besides, to prevent unfair comparisons from increased model parameters in downsampling operators (SpConv), we employ stride=1 SpConv for $d_{i}=1$ down and up operators. 
% %
% Comparing the results of $\{1,1,1\}$, $\{1,2,2\}$, and $\{1,2,4\}$, they have the same forward branches but differ in their backward branches' resolution.
% The larger down strides result in a large perception range.
% The progressively improved accuracy demonstrates the larger-scale receptive fields can effectively mitigate the loss of proximity.}
% it demonstrates that progressively increasing the downsampling rates in deeper networks can improve performance. 
% they have the same forward branches but differ in their backward branches' resolution.
% The results demonstrate that progressively increasing the downsampling rates in deeper networks can improve performance.
% Results for $\{1,1,1\}$, $\{1,2,2\}$, and $\{1,2,4\}$ demonstrate that progressively increasing the downsampling rates in deeper networks can improve performance. 
% In addition, we observe performance degradation with downsampling at the first stage. 

\textbf{Effectiveness of IWE.} Table~\ref{tab:ablations}(d) validates the capability of IWE to enhance spatial proximity. We compare IWE with some commonly used positional embedding methods~\cite{SST,DSVT} in 3D detection. Absolute position denotes the direct encoding of voxel coordinates using an MLP. The results demonstrate that IWE can significantly improve the detection performance by offering features with rich 3D positional and proximate information.

\begin{figure}[t]
\centering
\includegraphics[width=0.99\textwidth]{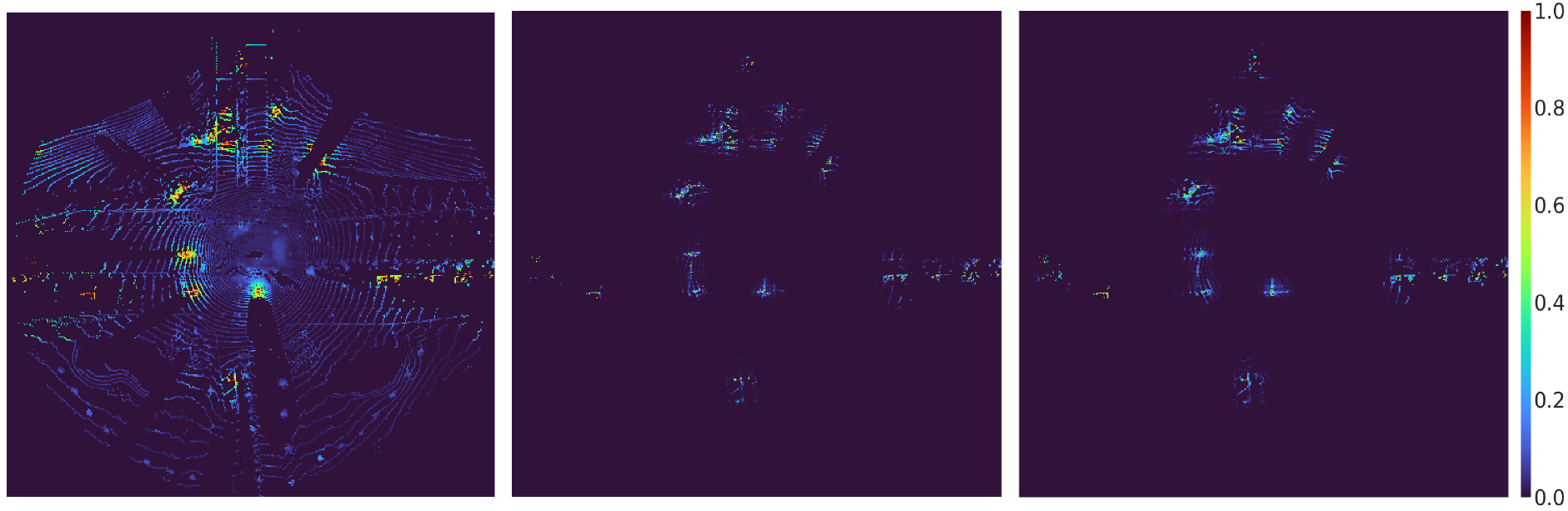}
\caption{The effective receptive fields (ERFs) of Voxel Mamba (left), group-based bidirectional Mamba (middle) and DSVT (right). }
\label{Fig::erf}
\vspace{-2mm}
\end{figure}

\subsection{Effective Receptive Field of Voxel Mamba}
Fig.~\ref{Fig::erf} illustrates the \textit{Effective Receptive Fields}~\cite{erf,replknet} (ERFs) of window partition-based method DSVT~\cite{DSVT}, group-based bidirectional Mamba and our proposed group-free method Voxel Mamba. 
For clear visualization, all models take pillars as inputs.
% Unlike 2D regular images, sparse voxels are not dense but rather scattered across the entire BEV map. 
The group partition in the group-based bidirectional Mamba is configured identically to DSVT.
Then, we randomly select voxels of interest from the ground truth bounding box and calculate the ERF at each non-empty voxel position. 
Subsequently, we merge the ERFs into a single image by taking the maximum value at each voxel location.
A wider activation area indicates a larger ERF.
From Fig.~\ref{Fig::erf}, we see that Voxel Mamba exhibits a notably larger ERF than DSVT and group-based bidirectional Mamba, which can be attributed to the benefits of group-free operation.
The larger ERF can cover a more complete local region and enhance the spatial proximity in 1D sequences.
% Besides, 

\label{sec:experiment}

\section{Conclusion}
\label{sec:conclusion}

In this paper, we proposed Voxel Mamba, a group-free SSM-based 3D backbone for point cloud based 3D detection. 
We first analyzed the proximity loss of group partition in current serialization-based 3D detection methods.
By taking the advantage of linear complexity of SSMs, we proposed a group-free strategy to alleviate the loss of spatial proximity in 3D to 1D serialization. We further proposed the DSB block and IWP strategy to build larger effective receptive fields and improve the spatial proximity of our Voxel Mamba framework.  
Experiments demonstrated that Voxel Mamba achieved state-of-the-art results on Waymo and nuScene datasets.
Without elaborated optimization, our model consumed less memory than group-based Voxel Transformer methods, and our group-free strategy was more efficient and deployment-friendly than group partition.
Voxel Mamba provided an efficient group-free solution for sparse point clouds for 3D tasks.

\textbf{Limitations.}
While the proposed Voxel Mamba achieves state-of-the-art performance in point cloud based 3D object detection, it still has some limitations to be further addressed. First, in the Hilbert Input Layer, the curve templates occupy a certain amount of GPU memory, which becomes more substantial as the voxel resolution increases. Besides, a more elaborately designed downsampling and upsampling operation could improve more the model efficiency. We will investigate these problems in future work.

% Voxel Mamba mainly focuses on 3D object detection in autonomous driving. We will investigate the application in indoor scenarios in future work

% \textbf{Limitations} Voxel Mamba mainly focuses on 3D object detection in autonomous driving. We will investigate the application in indoor scenarios in future work.

% \textbf{Broader impacts} Voxel Mamba can improve the safety of autonomous driving, but sometimes false predictions may lead to traffic accidents.
\label{sec:conclusion}

% \bibliographystyle{splncs04}
% {\small
% \bibliographystyle{ieee_fullname}
% \bibliography{egbib}
% }
% {\small
% \bibliographystyle{plain}
% \bibliography{egbib}
% }
\bibliographystyle{plain}
\bibliography{egbib}

% \bibliography{egbib}

\end{document}